\documentclass{article}
\usepackage[table,xcdraw]{xcolor}

\usepackage[preprint]{neurips_2022}

\usepackage[utf8]{inputenc} 
\usepackage[T1]{fontenc}    
\usepackage{hyperref}       
\usepackage{amsfonts}       
\usepackage{nicefrac}       
\usepackage{microtype}      

\usepackage{natbib} 
\usepackage{mathtools} 
\usepackage{booktabs} 
\usepackage{tikz} 
\newcommand{\comment}[1]{} 
\usepackage{graphicx}
\usepackage{amsmath}
\usepackage{amssymb}
\usepackage{url}
\usepackage{algorithmicx}
\usepackage{algpseudocode}
\usepackage{caption}
\usepackage{subcaption}
\usepackage{booktabs}
\usepackage{mathtools, xparse}
\usepackage{epstopdf}
\usepackage{algcompatible, amsmath}
\usepackage{commath}
\usepackage{float}
\usepackage{multirow}
\usepackage{pdflscape}
\usepackage{adjustbox}
\usepackage{bm}
\usepackage[ruled, lined, linesnumbered, commentsnumbered, longend]{algorithm2e}
\usepackage{authblk}
\usepackage{pdfpages}
\usepackage{etoolbox}
\usepackage{authblk}

\title{GCVAE: Generalized-Controllable Variational AutoEncoder}

\author[1]{Kenneth ~Ezukwoke\footnote{Corresponding author}}
\author[1]{Anis ~Hoayek}
\author[1]{Mireille ~Batton-Hubert}
\author[2]{Xavier ~Boucher}
\affil[1]{Department of Mathematics and Industrial Engineering}
\affil[2]{Department of Organisation and Environmental Engineering}
\affil[1, 2]{Mines Saint-Etienne, Univ Clermont Auvergne, INP Clermont Auvergne, CNRS, UMR 6158 LIMOS, F - 42023 Saint-Etienne, France \vskip 1mm
{\texttt{\{ifeanyi.ezukwoke, anis.hoayek, batton, boucher\}@emse.fr}}}

\comment{
\author{%
  Kenneth ~Ezukwoke\\
  Department of Mathematics and Industrial Engineering\\
  Cranberry-Lemon University\\
  Pittsburgh, PA 15213 \\
  \texttt{hippo@cs.cranberry-lemon.edu} \\
  \And
  Anis ~Hoayek\\
  Department of Mathematics and Industrial Engineering\\
  Cranberry-Lemon University\\
  Pittsburgh, PA 15213 \\
  \texttt{hippo@cs.cranberry-lemon.edu} \\
  \And
  Mireille ~Batton-Hubert\\
  Department of Mathematics and Industrial Engineering\\
  Cranberry-Lemon University\\
  Pittsburgh, PA 15213 \\
  \texttt{hippo@cs.cranberry-lemon.edu} \\
  \And
  Xavier ~Boucher\\
  Department of Organisation and Environmental Engineering\\
  Cranberry-Lemon University\\
  Pittsburgh, PA 15213 \\
  \texttt{hippo@cs.cranberry-lemon.edu} \\
}
}

\begin{document}

\maketitle

\begin{abstract}
Variational autoencoders (VAEs) have recently been used for unsupervised disentanglement learning of complex density distributions. Numerous variants exist to encourage disentanglement in latent space while improving reconstruction. However, none have simultaneously managed the trade-off between attaining extremely low reconstruction error and a high disentanglement score. We present a generalized framework to handle this challenge under constrained optimization and demonstrate that it outperforms state-of-the-art existing models as regards disentanglement while balancing reconstruction. We introduce three controllable Lagrangian hyperparameters to control reconstruction loss, KL divergence loss and correlation measure. We prove that maximizing information in the reconstruction network is equivalent to information maximization during amortized inference under reasonable assumptions and constraint relaxation.
\end{abstract}

\section{Introduction}
Unsupervised learning describes an intelligent approach for extracting meaningful classes of decisions from unlabeled data using intrinsic properties such as similarity metrics. Numerous classical data-driven approaches based on multivariate statistical analysis including Principal Component Analysis \citep{pearson, Hoffmann.2007, Song.2014} and Canonical Correlation Analysis \citep{hoteling} have been used to obtain latent or subspace patterns of high correlation. Decision labels are then obtained by applying clustering algorithms \citep{Martin.1996, lof, Andrade.2013,Liu.2019, Ren.2020}. Recently, deep learning-based approaches have gained traction for their improved performance at learning better latent representations with a high disentanglement metric \citep{hinton2006reducing} and have evolved into a new dimension of learning known as \textit{Disentangled Representation Learning} \citep{bengio2014representation}.

Disentanglement learning is a branch of unsupervised deep representation learning which involves uncovering the underlying independent factors responsible for making up an observation. For instance, an image with high dimensional features (eyes, nose, mouth, hair etc.), can be projected into a lower dimensional space of high disentanglement where different factored features of the images can be easily identified visually or clustered by an algorithm. Autoencoders are a specific type of a neural network, designed to encode the input into a compressed and meaningful representation, and then decode it back, so that the reconstructed input is as similar as possible to the original \citep{bank2021autoencoders}. While learning the reconstructed space, autoencoders are not capable of inducing disentanglement in the latent space, hence, keeping the latent factors entangled. 

Variational Autoencoders (VAE) \citep{kingma2014autoencoding} are an improved generative version of autoencoder based on probabilistic graphical modeling. VAEs replace the encoder section of an autoencoder with an amortized inference process (from Bayesian machine learning) and sample the latent space using a reparameterization trick before reconstruction.  VAEs have been employed for text disentanglement \citep{john2018disentangled} and generation \citep{shen2019generating}, as well as image generation \citep{kingma2014autoencoding, kulkarni2015deep, gregor2015draw}. Despite the numerous benefits of VAEs, they experience certain drawbacks including the notable vanishing Kullback-Leibler (KL) divergence problem. However, $\beta$-VAE \citep{betavae} solves the challenges arising from using VAE by imposing a weight ($\beta \gg 1$) on KL-divergence, thus eliminating the problem. This is at the expense of poor reconstruction which leaves room for improvement. InfoGAN \citep{infogan} belongs to a family of Generative Adversarial Networks (GAN) \citep{gan}, and involves an adversarial training of two models in a zero-sum game with the objective of inducing disentanglement in the latent space by maximizing variational mutual information. Despite the improved representation learned from using InfoGAN, a major drawback, as with all GANs, is the instability in training and strong assumption as regards the underlying distribution of the data.

It is imperative therefore to improve the amortized inference process of generative VAE networks which induces disentanglement with good reconstruction. InfoVAE \citep{zhao2018infovae} takes a cue from InfoGAN by introducing a weighted mutual information term within the original objective of the VAE. While it proves to be an improvement in disentanglement and representation compared to VAE and $\beta$-VAE, reconstruction is still not optimal. It is similar in performance to FactorVAE \citep{f_vae}, whose objective also includes mutual information and a directional KL-divergence in the latent dimension. FactorVAE uses a discriminating network to compare the performance of the latent space at every epoch, achieving better disentanglement compared to its counterparts. A challenging question to answer when  using FactorVAE is the choice of hyperparameter $\gamma$ capable of inducing maximum disentanglement in the latent space while balancing reconstruction. ControlVAE \citep{controlvae} on the other hand tries to solve the problem of selecting the appropriate $\beta$ automatically by introducing a non-linear Proportional-Integral-Derivative (\textbf{PID}) controller, to adaptively control the $\beta$ hyperparameter.

In all the methods proposed in the literature, very little has been done in trying to adapt a model that encourages disentanglement while minimizing reconstruction error in the VAE model simultaneously. We propose a generalized framework similar that emphasizes information maximization in the reconstruction network rather than in the inference network, and tune the hyperparameters using the non-linear PID controllers proposed in controlVAE \citep{controlvae}. The proposed model is designed to maximize the mutual information between the reconstructed data and the latent spaces by adaptive weighting of the reconstruction term of the VAE. We show that factorizing latent space or increasing the weight of $\beta$ on KL-divergence is not entirely sufficient for disentanglement. Our proposed model identifies the failures of existing models and introduces the squared Mahalanobis distance as a heuristic to improve independence in latent space.

The remaining sections of the paper are as follows: Section two discusses variational autoencoders and their variants; section three presents the proposed generalized-controllable variational autoencoder; section four contains experimental results; section five presents the final conclusion and finally the acknowledgements.

\section{Variational Autoencoder (VAE)}
VAEs \citep{kingma2014autoencoding} are a class of generative model proposed to model complex distributions existing in images, natural language and functional data. VAEs serve a significant purpose in language modeling, especially in representation learning for text disentanglement, unsupervised abstractive sentence summarization \citep{schumann2018unsupervised} and long and coherent text generation \citep{shen2019generating} among others. We formally define the VAE model by observing a $d$-dimensional input space $\{x_i\}_{i = 1}^{N} \in \mathcal{X}$ consisting of N-independently and identically distributed (\textbf{i.i.d}) samples; $k$-dimensional latent space $\{z_i\}_{i = 1}^{N} \in \mathcal{Z}$ (where $k \ll d$)  over which a generative model is defined. We assume an empirical prior distribution $p_{\theta}(z) \sim \mathcal{N}(0, I)$ to infer an approximate posterior distribution $q_{\phi}(z|x) \sim \mathcal{N}(z|\mu_{\phi}(x), \sigma_{\phi}^{2}(x)I)$, with mean $\mu_{\phi}(x)$ and variance $\sigma_{\phi}^{2}(x)I$ used for reparameterization sampling of the latent space $z$ \citep{kingma2014autoencoding}. We model the data using conditional distribution $p_{\theta}(x|z) \sim \mathcal{N}(x|\mu_{\theta}(x), \sigma_{\theta}^{2}(x)I)$. Let us suppose that the underlying distribution of the input space $p_(x)$ follows a normal distribution, and its empirical distribution is denoted by $p_{\mathcal{D}}(x)$. The objective function $\mathcal{L}(\theta, \phi)$ to be maximized is given by,
\small
\begin{align}
     \underset{p_{\mathcal{D}}}{\mathbb{E}} \underset{z \sim q_{\phi}(z|x)}{\mathbb{E}}[\ln p_{\theta}(x|z)]
    -   D_{KL}(q_{\phi}(z|x)|| p_{\theta}(z)) \label{maxmin}
\end{align}
\normalsize
where $\phi$ and $\theta$ represent the parameters of the neural network encoder and decoder respectively.
This objective function is known as the \textbf{E}vidence \textbf{L}ower \textbf{Bo}und (\textbf{ELBO}), upper bounded by the input data log-likelihood $\ln p_{\theta}(x)$. The first term of the objective is a \textit{reconstruction error}. Ideally, maximizing negative reconstruction loss, simultaneously minimizes non-symmetric divergence between the distributions of the \textit{Kullback-Leibler (KL)-divergence} terms. 

Equivalently, maximizing ELBO maximizes negative KL-divergence between the inference prior $q_{\phi}(z|x)$ and generative prior $p_{\theta}(z|x)$, where $p_{\theta}(z|x) \propto p_{\theta}(x, z)$ and $q_{\phi}(z|x) \propto q_{\phi}(x, z)$.

\small
\begin{align}
     \mathcal{L}(\theta, \phi) &= - \underset{p_{\mathcal{D}}}{\mathbb{E}}[D_{KL}(q_{\phi}(z|x)|| p_{\theta}(z|x))] \\
     \ln p_{\theta}(x) &\geq \underset{p_{\mathcal{D}}}{\mathbb{E}} \left[\ln p_{\theta}(x|z) - D_{KL}(q_{\phi}(z|x)||p_{\theta}(z))\right] \label{alt1}
\end{align}
\normalsize
We prove equation (\ref{alt1}) in Appendix \ref{neg_kl}.\\
\textbf{$\beta$-VAE} \citep{betavae}
introduces a fixed weight on the KL divergence ($\beta D_{KL}$) with the aim of (i) eliminating the vanishing KL problem (where $D_{KL} \rightarrow 0$) and (ii) inducing disentanglement by adjusting the value of $\beta$.

\subsection{ControlVAE}
Due to the difficulty in modeling complex entangled functions and the lack of statistical independence of the latent variables, \citep{betavae} has proposed an adjustable $\beta$-hyperparameter that (i) balances latent bottlenecks and independence with an improved reconstruction compared to VAE (ii) solves the vanishing KL-divergence problem where $D_{KL}(q_{\phi}(z|x)|| p_{\theta}(z)) = 0$. 
\small
\begin{align}
     \underset{p_{\mathcal{D}}}{\mathbb{E}} \underset{z \sim q_{\phi}(z|x)}{\mathbb{E}}[\ln p_{\theta}(x|z)]
    -   \beta D_{KL}(q_{\phi}(z|x)|| p_{\theta}(z))
\end{align}
\normalsize
$\beta$-VAE ensures that the hyperparameter on the KL increases the independence of the latent variables for very large values of $\beta$ ($\beta > 1$) while ensuring posterior conditional inference probability $q_{\phi}(z|x)$ is relatively close to its prior $p_{\theta}(z)$. However, there is no a priori limit for $\beta$ as the value of $\beta$ depends on the data; which results in overfitting of the prior to the approximate posterior $p_{\theta}(z) \approx q_{\phi}(z|x)$. This implies that the information about $x$ preserved in $z$ decreases with $\beta \rightarrow +\infty$. \textbf{ControlVAE} \citep{controlvae} proposes the use of a PID controller for an adaptive $\beta$. Its objective function is as follows,
\small
\begin{align}
     \underset{p_{\mathcal{D}}}{\mathbb{E}} \underset{z \sim q_{\phi}(z|x)}{\mathbb{E}}[\ln p_{\theta}(x|z)]
    -   \beta_t D_{KL}(q_{\phi}(z|x)|| p_{\theta}(z))
\end{align}
\normalsize
Where,
\small
\begin{align}
     \beta_t = \frac{K_p}{1+ exp(e_t)} - K_i \sum_{j = 0}^{t}e_j + \beta_{min}
\end{align}
\normalsize
where $K_p$ and $K_i$ are constants; $e_t$ is the error between the actual KL-divergence value and the desired value at time $t$; $\beta_{min}$ is an application-specific constant which shifts the range to within which $\beta_t$ is allowed to vary.

\comment{
\subsection{FactorVAE}
FactorVAE \citep{f_vae} auguments the VAE objective with a term that directly encourages independence in latent space, by further decomposing the $D_{KL}$ term into two terms including the mutual information between $x$ and $z$ and the total correlation. The objective function for factorVAE is given by,

\small
\begin{align}
     \underset{p_{\mathcal{D}}}{\mathbb{E}} &\underset{z \sim q_{\phi}(z|x)}{\mathbb{E}}[\ln p_{\theta}(x|z)]
    - D_{KL}(q_{\phi}(z|x)|| p_{\theta}(z)) \notag\\
     &\qquad \quad - \gamma D_{KL} (q_{\phi}(z)||\bar{q}_{\phi}(z)) \label{factorvae}
\end{align}
\normalsize
Where $\gamma$ is the hyperparameter to optimize and $\bar{q}_{\phi}(z)=  \Pi_{j = 1}^{d}q(z_j)$ is the discriminating latent variable. The last term of (\ref{factorvae}) is called the total correlation term.
}

\subsection{InfoVAE}
InfoVAE \citep{zhao2018infovae} adds a weighted mutual information criteria to the VAE objective function and maximizes the mutual information between the data $x$ and latent space $z$.
The objective function of InfoVAE is given by,
\small
\begin{align}
     \underset{p_{\mathcal{D}}}{\mathbb{E}} \underset{z \sim q_{\phi}(z|x)}{\mathbb{E}}[\ln p_{\theta}(x|z)]
    - (1-\alpha) D_{KL}(q_{\phi}(z|x)|| p_{\theta}(z)) - (\alpha + \lambda - 1)D_{KL} (q_{\phi}(z)||p_{\theta}(z))
\end{align}
\normalsize
which is reduced to a family of variational autoencoding objectives when $\alpha = 1$,
\small
\begin{align}
     \underset{p_{\mathcal{D}}}{\mathbb{E}} &\underset{z \sim q_{\phi}(z|x)}{\mathbb{E}}[\ln p_{\theta}(x|z)]
     - \lambda D_{KL} (q_{\phi}(z)||p_{\theta}(z)) \label{infovae}
\end{align}
\normalsize

Where $D_{KL} (q_{\phi}(z)||p_{\theta}(z))$ is a non-negative distance called Maximum Mean Discrepancy (\textbf{MMD}) and is defined as a kernel function,
\small
\begin{align}
     D^{2}_{MMD}(q||p) = \underset{p(z)\cdot p(z^{\prime})}{\mathbb{E}}[k(z, z^{\prime})] - \underset{q(z)\cdot p(z^{\prime})}{2\mathbb{E}}[k(z, z^{\prime})] + \underset{q(z)\cdot q(z^{\prime})}{\mathbb{E}}[k(z, z^{\prime})] \label{mmd}
\end{align}
\normalsize
In Equation (\ref{mmd}), $k(z, z^{\prime})$ represents a positive semi-definite kernel matrix \citep{book_svm}. $D^{2}_{MMD}(q||p)$ measures the distance between the sample distribution $z$ and true distribution $z^{\prime} \sim \mathcal{N}(0, I)$ and $\lambda > 0$ is a scaling coefficeint \citep{zhao2018infovae}.

\section{Generalized-Controllable VAE} \label{gcvae:sec}
\subsection{Alternate formulation of ELBO, $\beta$-VAE and InfoVAE objective} \label{altelbo}
We consider an alternate formulation of the InfoVAE objective based on mutual information maximization. Let $I_p(x^{\prime}, z)$ be the mutual information between the reconstructed data $x^{\prime}$ and $z$ under joint distribution $p_{\theta}(x^{\prime}|z)p_{\theta}(z)$, where $p_{\theta}(z) \sim \mathcal{N}(0, I)$.  $I_q(x, z)$ is the mutual information between $x$ and $z$ under joint distribution $q_{\phi}(z|x)p_{\mathcal{D}}(x)$. Since the objective of the VAE is to minimize a certain reconstruction loss, and considering that this loss is an $l_2$-norm, we assume that the expected log-likelihood  $\mathbb{E}[\ln p_{\theta}(x^{\prime}|z)]$ is approximately equal to $\mathbb{E}[\ln p_{\theta}(x|z)]$ and $p_{\mathcal{D}}(x^{\prime}) \approx p_{\mathcal{D}}(x)$ at the optimal point for $n$-number of experiments.

\textbf{Proposition 1.} Let $x \in \mathcal{X}$ and $z \in \mathcal{Z}$ be random variables with $q_{\phi}(x)$ and $q_{\phi}(z)$ corresponding to their marginal probability density distribution. We denote the  joint probability density function between $x$ and $z$ as $q_{\phi}(x, z)$ where $q_{\phi}(z) \sim \mathcal{N}(\mu_{\phi}, \sigma_{\phi}^{2} I)$ and $p_{\theta}(x) \sim \mathcal{N}(\mu_{\theta}, \sigma_{\theta}^{2} I)$. For known parameters of $x$ and $z$, the mutual information between $x$ and $z$ for the inference network is expressed as
\small
\begin{align}
     I_q(x, z) &= \int_x \int_z q_{\phi}(x,z) \ln \frac{q_{\phi}(x,z)}{q_{\phi}(z)q_{\phi}(x)} \, dx \, dz = D_{KL}(q_{\phi}(x,z)p_{\theta}(z)||q_{\phi}(z)q_{\phi}(x)p_{\theta}(z)) \label{i_q}\\
     &= D_{KL}(q_{\phi}(z|x)||p_{\theta}(z)) - D_{KL}(q_{\phi}(z)||p_{\theta}(z))
\end{align}
\normalsize
Furthermore, we can reformulate Equation (\ref{i_q}) by introducing the joint generative distribution so that,
\small
\begin{align}
     I_q(x, z) &= D_{KL}(q_{\phi}(z|x)p_{\theta}(x, z)||q_{\phi}(z)p_{\theta}(x, z)) \label{i_qq}\\
         &= D_{KL}(q_{\phi}(z|x)||p_{\theta}(z)) - D_{KL}(q_{\phi}(z)||p_{\theta}(z)) - \underset{z \sim q_{\phi}(z|x)}{\mathbb{E}}[\ln p_{\theta}(x|z)] + \ln p_{\theta}(x)
\end{align}
\normalsize
The complete derivation is found in the Appendix (section \ref{miq}). This maximization is also known to improve reconstruction of generative models, as seen for GANs \citep{pmlr_v80_belghazi18a} while the minimized term ensures an effective information bottleneck. The constraint optimization formulation used here will be applied in a later section.

\begin{figure}[ht]
    \centering
    \includegraphics[scale= 0.17]{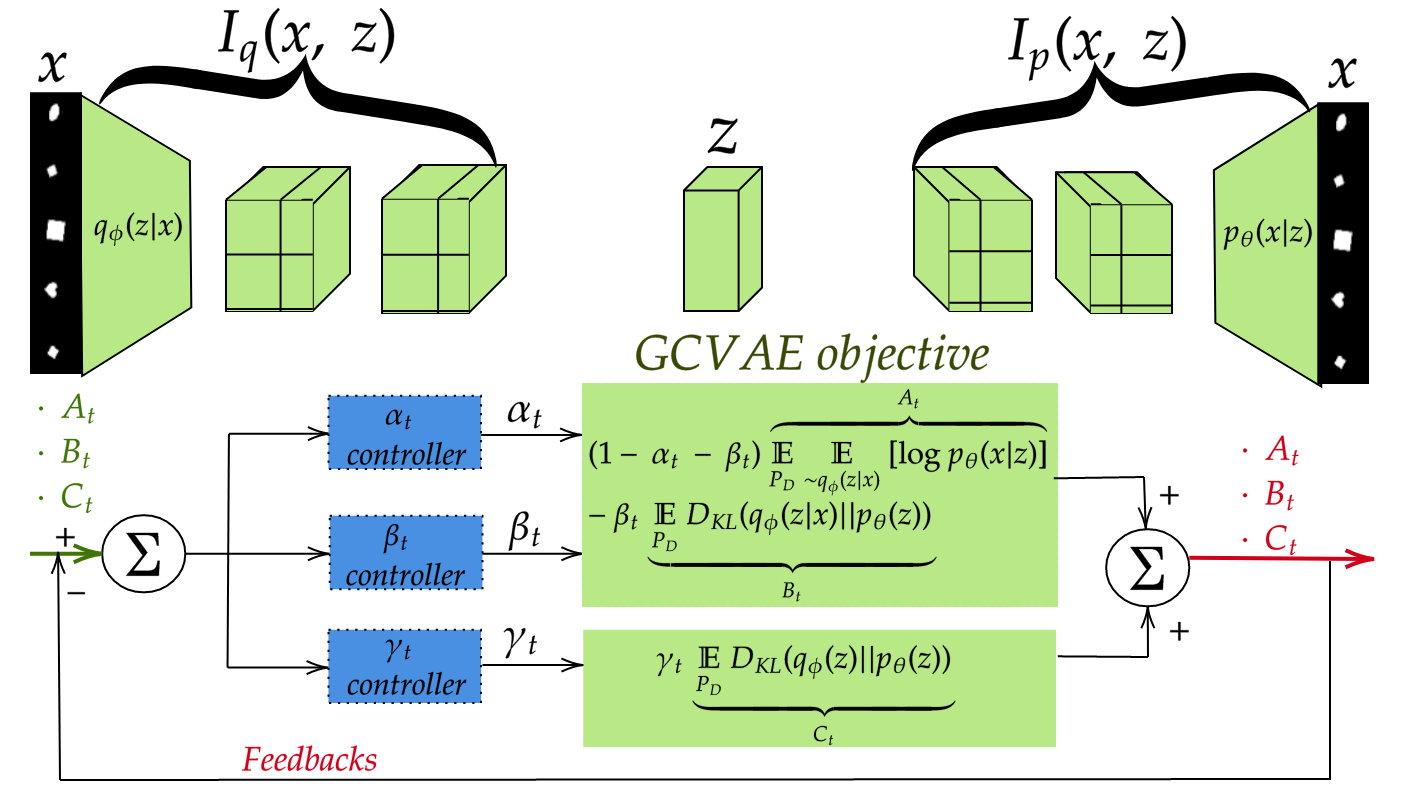}
     \caption{\label{gcvae_img} GCVAE framework. $\alpha_t, \beta_t$ and $\gamma_t$ respectively provide automatic balancing of the log-likelihood and KL divergences for optimal reconstruction and disentanglement. The feed-ins $A_t, B_t$ and $C_t$ are expectations of variational loss.}
\end{figure}

Mutual information from the generative network is given as,
\small
\begin{align}
     I_p(x^{\prime}, z) &= D_{KL}(p_{\theta}(x^{\prime},z)||p_{\theta}(z)p_{\theta}(x^{\prime})) = \underset{z \sim q_{\phi}(z|x)}{\mathbb{E}}[\ln p_{\theta}(x^{\prime}|z)] - \ln p_{\theta}(x^{\prime}) \label{eqn:fourteen}
\end{align}
\normalsize
The marginal log-likelihood $\ln p_{\theta}(x^{\prime})$, given as $\int_z p_{\theta}(x^{\prime}|z)p_{\theta}(z) dz$ with an integral over $z$ is computationally intractable for a large number of $z$-variables.  We can drop the marginal since it does not significantly impact the likelihood term. We reformulate the objective function of a family of InfoVAE (Equation \ref{infovae}) as a function of the mutual information of both generative and reconstruction processes as follows,
\small
\begin{align}
     \underset{\phi, \theta}{max} \quad I_p(x^{\prime}, z) - I_q(x, z) \label{reforminfo}
\end{align}
\normalsize
We call this formulation (Equation \ref{reforminfo}), the total mutual information of a VAE, which is further explained in Figure \ref{gcvae_img}. This implies therefore, that accurately reconstructing the original distribution $p_{\mathcal{D}}(x)$ requires us to maximize the mutual information $I_p(x^{\prime}, z)$ in the reconstructed space while reducing information loss $I_q(x,z)$ during inference. The constraint optimization formulation of Equation (\ref{reforminfo}) can be written as,
\small
\begin{align}
     \underset{\phi, \theta}{max} \quad &I_p(x^{\prime}, z) \notag \\
     s.t \qquad & D_{KL}(q_{\phi}(z|x)||p_{\theta}(z)) \leq \xi_1 \notag\\
     s.t \qquad & D_{KL}(q_{\phi}(z)||p_{\theta}(z)) \leq \xi_2 \notag \\
     & \qquad \xi_1, \xi_2 \geq 0 \label{reforminfo1}
\end{align}
\normalsize
$\xi_1$ and $\xi_2$ in (\ref{reforminfo1}) are permissible error terms. We apply the Karush–Kuhn–Tucker (KKT) conditions \citep{karush39minima, kuhn1951nonlinear} and solve for optimality, by applying the Lagrangian multipliers to obtain the following,
\small
\begin{align}
     \underset{\phi, \theta}{max} & \quad \mathcal{L}(\phi, \theta, \lambda, \beta)\\
     \underset{\phi, \theta}{max} &\underset{z \sim q_{\phi}(z|x)}{\mathbb{E}}[\ln p_{\theta}(x|z)] - \beta D_{KL}(q_{\phi}(z|x)||p_{\theta}(z)) - \lambda D_{KL}(q_{\phi}(z)||p_{\theta}(z))
\end{align}
\normalsize
\small
\begin{equation}
    \mathcal{L}(\phi, \theta, \lambda, \beta) = 
    \begin{cases}
    ELBO & \text{if } \beta=1, \lambda = 0\\
    \beta-VAE & \text{if } \beta>1, \lambda = 0\\
    InfoVAE & \text{if } \beta=0, \lambda \geq 1\\
    FactorVAE & \text{if } \beta=1, \lambda = -1
\end{cases}
\end{equation}
\normalsize
We derive from Equation (\ref{reforminfo}), that maximizing mutual information under reasonable constraints in the generative network of VAE is equivalent to mutual information minimization in the inference network. Hence, maximizing $I_p(x^{\prime}, z)$ subject to inference constraints is equivalent to maximizing the total mutual information. The constraints to which mutual information is subjected in the InfoVAE optimization framework does not simultaneously balance the high disentanglement-low reconstruction error trade-off. In the next section, we propose an optimization framework based on information maximization in a generative network, that simultaneously manages the trade-off between high disentanglement and loss reconstruction error.

\subsection{GCVAE: Generalized-Controllable VAE}

We propose a generalized framework for variational inference modeling, taking into account the idea of information maximization in the reconstruction network. We prioritize the disentanglement of the latent space and balancing the trade-off between disentanglement metric and reconstruction loss by maximizing the mutual information between that reconstructed data $x^{\prime}$ and latent space $z$. Furthermore, we show that weighting the reconstruction term of this new lower bound by the controllable weight of its KL-divergence significantly improves the disentangling factor. Under the constraint optimization framework, we follow a similar reformulation, as shown in Equation (\ref{reforminfo1}), to propose the following formal problem.

\textbf{Proposition 2.} Let $x \in \mathcal{X}$ and $z \in \mathcal{Z}$ be continuous spaces representing the input and latent space respectively. $I_p(x^{\prime}, z)$ is the joint mutual information space between $x^{\prime}$ and $z$ generated from the posterior $p_{\theta}(x|z)$ after obtaining an inference posterior $q_{\phi}(z|x)$. Let $0 \leq \beta_t \leq \alpha_t \leq 1$ be controllable optimizable hyperparameters (Lagrangian multipliers). We reach an optimum when $q_{\phi}(z|x) \approx p_{\theta}(z|x)$. We maximize the mutual information in the reconstruction space subject to inference constraints as follows,
\small
\begin{align}
    & \underset{\theta, \phi, \xi^{+}, \xi^{-}, \xi_{p} \in \mathbb{R}}{max}\quad I_p(x^{\prime}, z) \notag\\
    s.t &\quad  \underset{p_{\mathcal{D}}}{\mathbb{E}} D_{KL}(q_{\phi}(z|x)\parallel p_{\theta}(z)) + I_p(x^{\prime}, z) \leq \xi^{-}\notag\\ 
    s.t &\quad - \underset{p_{\mathcal{D}}}{\mathbb{E}} D_{KL}(q_{\phi}(z)||p_{\theta}(z))\leq \xi^{+} \notag\\
    s.t &\quad I_p(x^{\prime}, z) \leq \xi_{p} \notag\\
    s.t &\quad \xi_i^{+}, \xi_i^{-}, \xi_{ip} \geq 0, \quad \forall i = 1, \dots, n
\end{align}
\normalsize
We apply the KKT conditions and solve for optimality as previously seen in section \ref{altelbo}  (proof provided in the Appendix \ref{gcvae}). Hence, 
\small
\begin{align}
    \mathcal{L}(\theta, \phi, \xi^{+}, \xi^{-}, \xi_p, \alpha, \beta, \gamma) &= (1-\alpha_t-\beta_t)\underset{p_{\mathcal{D}}}{\mathbb{E}} \underset{z \sim q_{\phi}(z|x)}{E}[\ln p_{\theta}(x|z)] - \beta_t \underset{p_{\mathcal{D}}}{\mathbb{E}} D_{KL}(q_{\phi}(z|x)||p_{\theta}(z)) \notag\\
    &\quad + \gamma_t \underset{p_{\mathcal{D}}}{\mathbb{E}} D_{KL}(q_{\phi}(z)||p_{\theta}(z))
    \label{gcvae:eqn1}
\end{align}
\normalsize
$\alpha_t$, $\beta_t$ and $\gamma_t$ and Lagrangian multipliers and are used here as controllable hyperparameters. The controllable weight on the reconstruction loss $1- \alpha_t - \beta_t$, implies the effect of the KL-divergence loss on the reconstruction loss, which simultaneously control the disentanglement in the latent space with respect to the reconstruction quality.
Empirical results show that this new lower bound is free from the vanishing KL-divergence problem even for fixed $\beta$ ($\beta \gg 1$) and does not overfit during training. Exponentially small values of $\alpha, \beta \rightarrow 0$ penalize reconstruction error and KL divergence respectively. The last term in (\ref{gcvae:eqn1}), $D_{KL}(q_{\phi}(z)||p_{\theta}(z))$ is the expected Mahalanobis distance (MD) denoted as $\mathbb{E} D_{MAH}^{2} (q||p)$ between the density function of two continuous variables $p$ and $q$ as opposed to MMD used in InfoVAE. MD is a positive semi-definite distance and obeys the triangle inequality. We show that MD distance is a better disentangling metric than MMD with better reconstruction quality.

\textbf{Proposition 3.} Let $x \in \mathcal{X}$ be independent random variables with distribution $p(x)$ and similarly $y \in \mathcal{Y} \sim q(y)$. $x^{\prime}$ and $y^{\prime}$ are transposes of $x$ and $y$ respectively.
We suppose that $x$ and $y$ are mapped to a Reproducing Kernel Hilbert Space (RKHS) denoted as $\mathcal{H}$ \citep{rkhs_1} so that $x \sim \varphi(x)$ and $y \sim \varphi(y)$ with a common covariance matrix $\Sigma_{d\times d}$. Let $z \coloneqq \{x_i, y_i\}_{i = 1}^{n} \in \mathcal{Z}$ be $n$ i.i.d random samples drawn from a distribution $\mathcal{D}_{z}$. The expected squared Mahalanobis distance between distributions $p$ and $q$ in a reproducing kernel Hilbert space is given by $\Sigma^{-1} D_{MMD}^{2}(q||p)$, where $\Sigma^{-1}$ is the diagonal covariance.
\small
\begin{align}
    \mathbb{E} D_{MAH}^{2} (q||p) &= \mathbb{E}[(\varphi(x) - \varphi(y))^{\prime}\Sigma^{-1}(\varphi(x) - \varphi(y))] \notag\\
    &\geq \mathbb{E}[tr(\Sigma^{-1} (\varphi(x) - \varphi(y))^{\prime}(\varphi(x) - \varphi(y)))] \notag\\
    &\geq tr\left(\parallel \underset{x\sim p}{\mathbb{E}}\varphi(x) - \underset{y\sim q}{\mathbb{E}}\varphi(y) \parallel^{2}_{\Sigma^{-1}, \mathcal{H}}\right) \notag\\
    &\geq tr(\Sigma^{-1} D_{MMD}^{2}(q||p))\notag\\
    \label{20}
\end{align}
\normalsize
\comment{
\begin{figure*}[ht]
\centering
    \begin{tabular}{ccccc}
    \subfloat[PCA]{\includegraphics[width=80px,height=2.5cm]{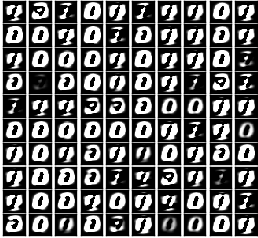}}\vspace{0.00mm}
    \subfloat[VAE]{\includegraphics[width=80px,height=2.5cm]{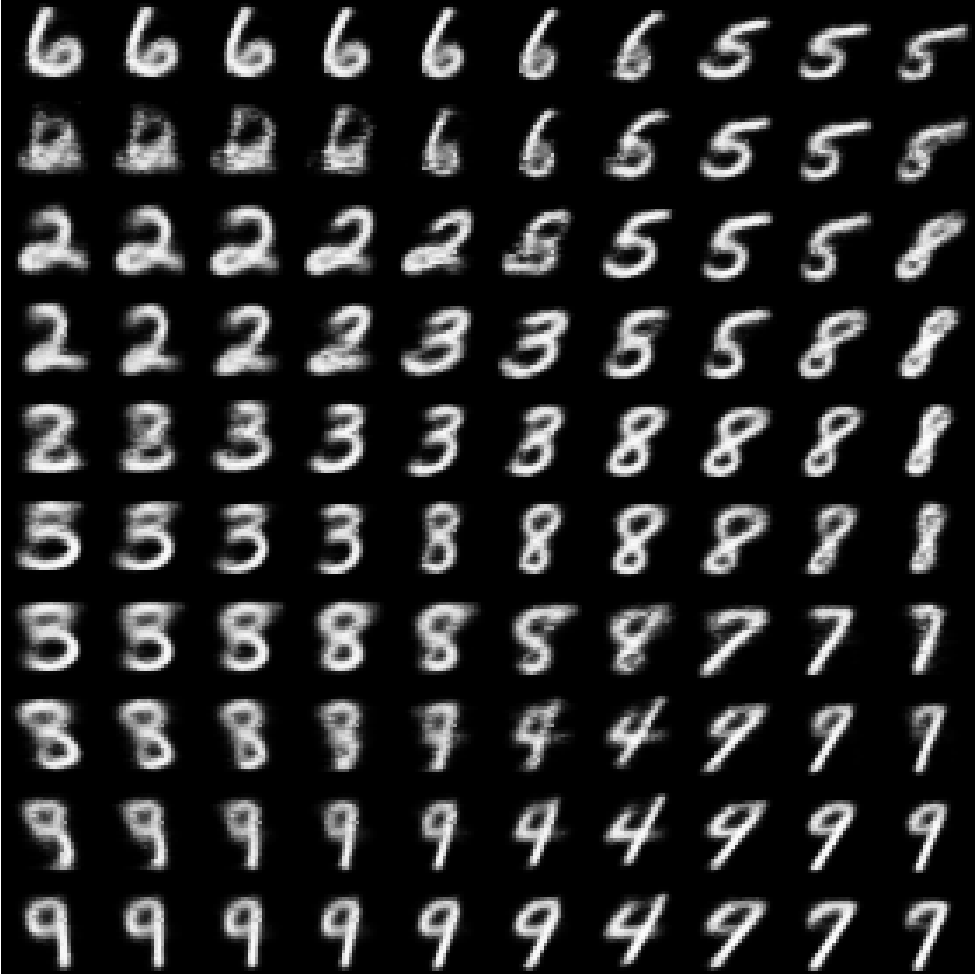}}\vspace{0.00mm}
    \subfloat[ControlVAE]{\includegraphics[width=80px,height=2.5cm]{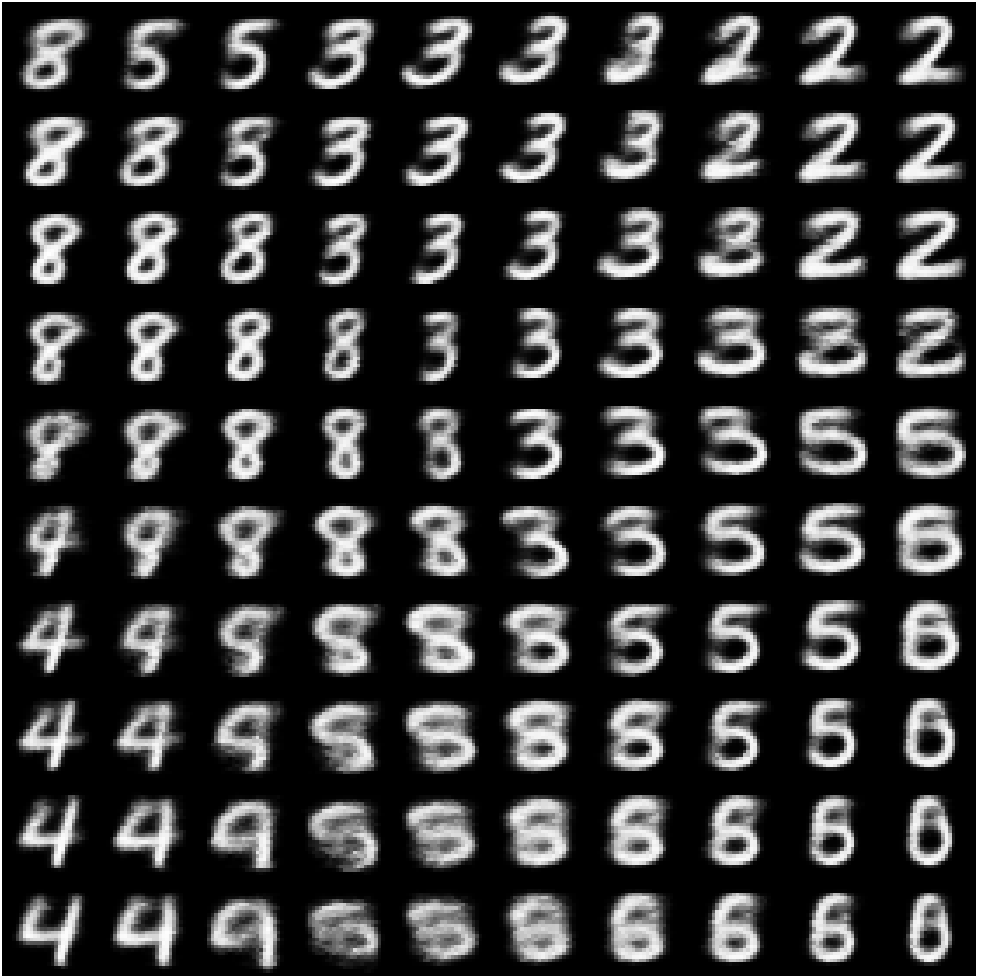}}\vspace{0.00mm}
    \subfloat[InfoVAE]{\includegraphics[width=80px,height=2.5cm]{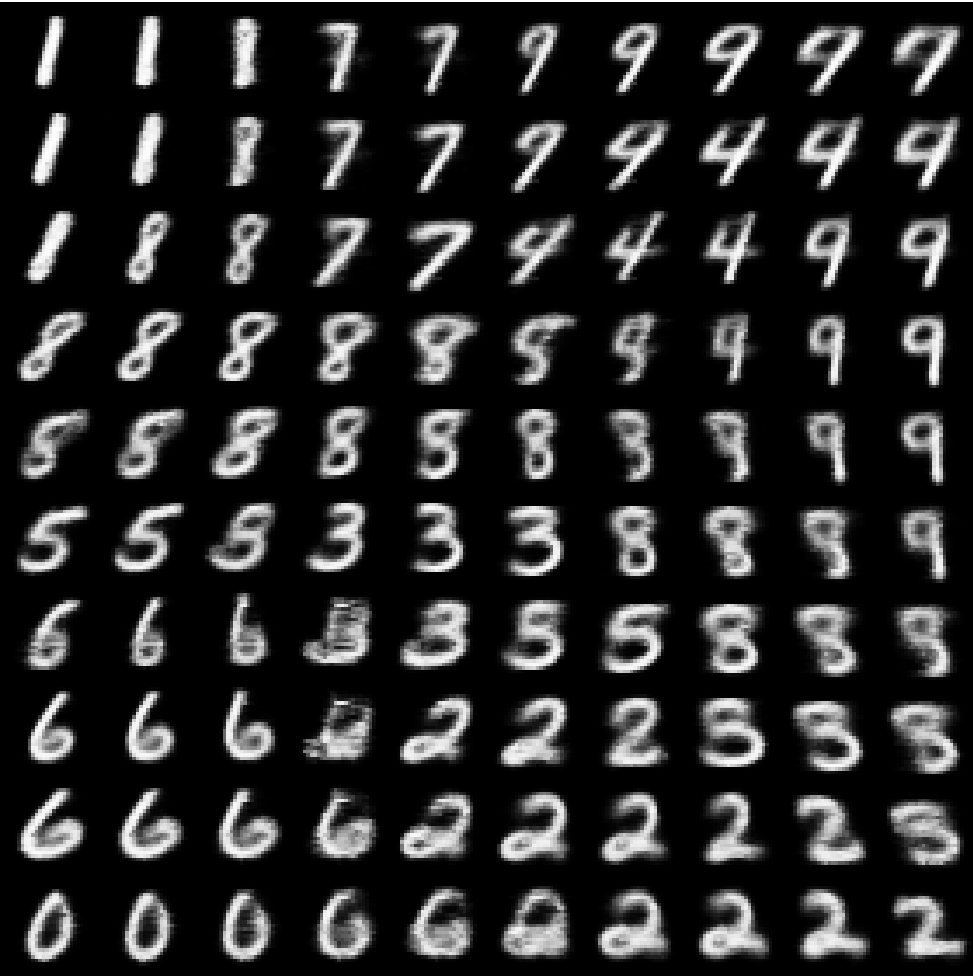}}\vspace{0.00mm}
    \subfloat[GCVAE]{\includegraphics[width=80px,height=2.5cm]{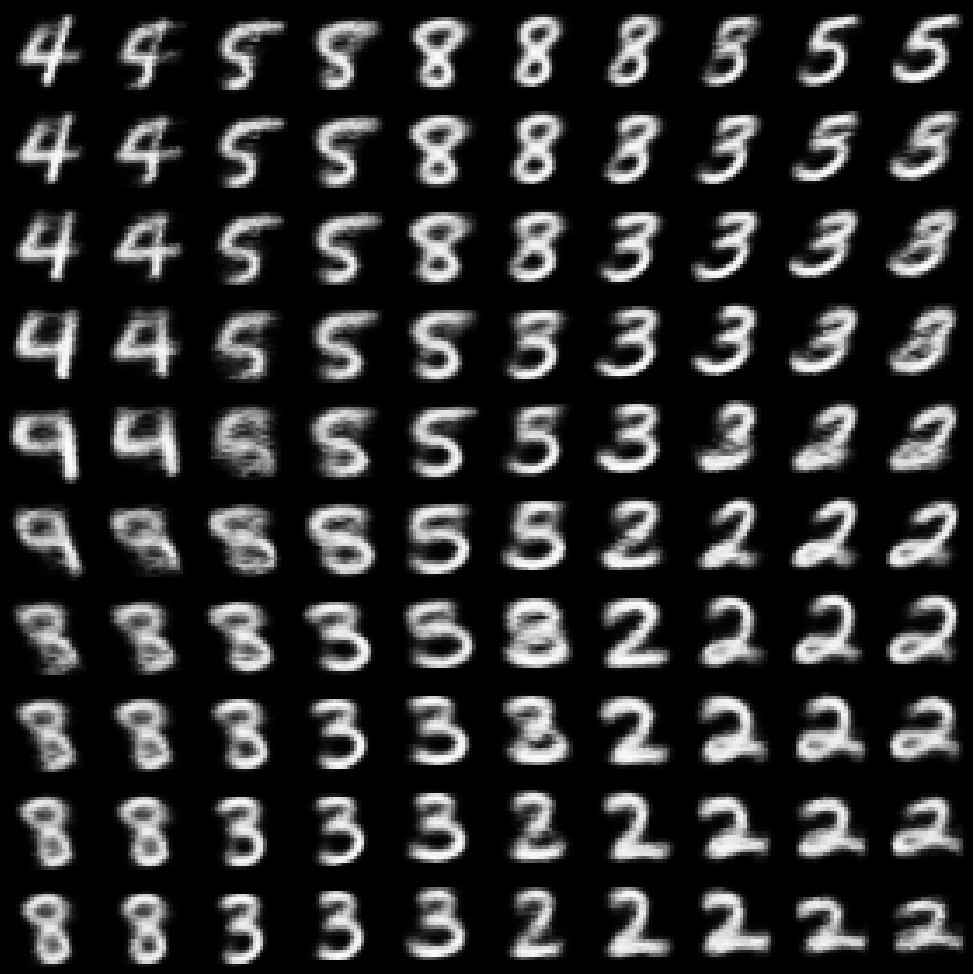}}
    \end{tabular}
  \caption{Reconstruction of the MNIST for PCA, VAE, ControlVAE, Info and GCVAE respectively. PCA representation is uninterpretable despite having high disentanglement.}
\end{figure*}
}

The expectation of the right is $\geq tr(\mathbb{E}\Sigma^{-1} D_{MMD}^{2}(q||p))$. The proof of Equation (\ref{20}) is given in Appendix \ref{mmd_mal_prove}. $\mathbb{E} D_{MAH}^{2} (q||p)$ is a measure of the average dissimilarity between the distributions $p$ and $q$ in the Hilbert space, as it normalizes the similarity with feature variances, therefore encouraging class discrimination. $\mathbb{E} D_{MAH}^{2} (q||p)$ reduces to $D_{MMD}^{2}(q||p)$ when $\Sigma^{-1}$ is identity. We note that $x$ is zero-centered before it is projected into the Hilbert space.\\
We deduce other variants of VAE from GCVAE loss as follows,
\small
\begin{equation}
\mathcal{L}(\theta, \phi, \xi^{+}, \xi^{-}, \xi, \alpha, \beta, \gamma) =
   \begin{cases}
        ELBO & \text{if } \alpha_t = \alpha = -1, \beta_t = \beta = 1, \gamma = 0\\
        ControlVAE & \text{if } \alpha_t = \alpha = 0, \beta_t > 0, \gamma = 0\\
        InfoVAE & \text{if } \alpha_t = \alpha = 0, \beta_t = \beta = 0, \gamma_t >1\\
        FactorVAE & \text{if } \alpha_t = \alpha = -1, \beta_t = \beta = 1, \gamma =-1
    \end{cases}
\end{equation}
\normalsize
While the above models focus on penalizing KL-divergence to induce disentanglement, GCVAE takes a parameter into account to control reconstruction loss ($\alpha_t$) while gradually improving disentanglement. GCVAE achieves optimality for $\alpha_t \in [0, 1-\beta_t], \beta_t \in [0, 1- \alpha_t], \gamma_t \in [0, 1]$. This new objective function allows us to significantly control the amount of meaningful information encoded in the latent space. 

\section{Experiment}
\subsection{Experimental setup and parameter optimization}
We design and select an experimental setup that places emphasis on the relevance of disentanglement without overlooking the issue of reconstruction trade-off encountered by some state-of-the art models. We validate the performance of the proposed model both qualitatively (reconstruction) and quantitatively (disentanglement), using a simplified version of \citep{f_vae} neural architecture. We train for $250K$ epochs with batch of 64 and a learning rate of $1e^{-3}$. The number of dimensions of the latent $z$ is $10$ for the DSprites dataset and $6$D for the 3D Shapes dataset. We prefer dimension $2$ to analyze the reconstruction and generative quality of the model. In all cases, the Adam optimizer is used for training the models.\\
\textbf{Model architecture}. The preferred model is a 4-layer, 2D convolutional neural network with (2, 2) filters. A full description of the network architecture is available in Appendix \ref{archi} of this paper.\\
\textbf{Parameter optimization}. The GCVAE lower bound can be reduced to the ELBO of other variational models, while the choice of values for the hyperparameters can be selected according to the original papers for the training phase. InfoVAE only enables the $\lambda$ weight on the MMD to be optimized ($\lambda  = 1000$ is used in our experiment). The value of $\beta=10$ is used for training $\beta$-VAE. ControlVAE optimizes the $\beta$ and we set an expected $\beta = 10$ for training. However, the GCVAE model requires the optimization of three PID-controllable parameters $\alpha_t$, $\beta_t$ and $\gamma_t$. The expected $\alpha$, $\beta$ and $\gamma$ hyperparameters for GCVAE are set to $10, 30, 0.1$ before training. These set points are selected after extensive trial and errors. To evaluate the strength of disentanglement and the quality of reconstruction, we propose three families of GCVAE according to the metric selected for the $D_{KL}(q_{\phi}(z)||p_{\theta}(z))$:\\ 
(1) \small \textbf{GCVAE-I}: $D_{KL}(q_{\phi}(z)||p_{\theta}(z))$ $=$ $D_{MMD}^{2}(q||p)$;\\
(2) \small \textbf{GCVAE-II}: $D_{KL}(q_{\phi}(z)||p_{\theta}(z))$ $=$ $D^2_{MAH}(q||p)$\\
(3) \small \textbf{GCVAE-III}: $D_{KL}(q_{\phi}(z)||p_{\theta}(z))$ $=$ $\mathbb{E} \Sigma^{-1} D_{MMD}^{2}(q||p)$.
\normalsize
\subsection{Dataset}
\textbf{MNIST} \citep{deng2012mnist} is a $28 \times 28$-dimensional image data set with $70,000$ observation. The data set is split into training observations ($60,000$) and testing observations ($10,000$) and contains images of numbers between $0-9$ values together with their labels.\\
\textbf{DSprites 2D Shapes}
DSprites \citep{dsprites17} is a data set of 2D $64 \times 64$-dimensional image data set with 737,280 binary observations. Five ground truth factors are available with properties (shape: 3; scale: 6; orientation: 40; x-position: 32; y-position: 32).
\subsection{Evaluation metric}
\comment{
\begin{figure}[ht]
    \centering
    \includegraphics[scale = 0.45]{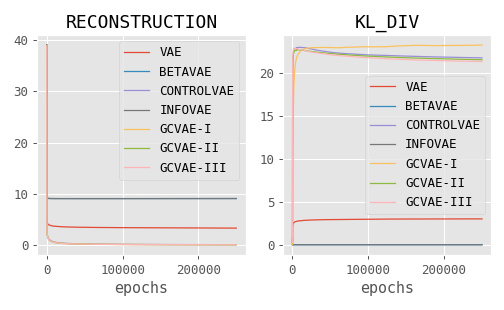}
     \caption{Reconstruction loss and KL divergence for DSprites data after training on 737 samples.}
\end{figure}
}
\comment{
\textbf{FactorVAE Score} \citep{f_vae} or \textbf{Z-min variance score}. Measures the accuracy of majority vote classifier that predicts the index of a fixed ground-truth factor ($y_k$) using latent code dimensions ($z_j$'s). It is computed by normalizing the latent factor by their standard deviation ($\sigma$). A random sample of the $y_k$ is selected and compared with $z_j$ and their variance computed. The code with the minimum variance best predict the fixed factor over a majority vote classifier for several subset of the fixed factors. 
}
\textbf{Mutual Information Gap (MIG) score} \citep{chen2019isolating}. Normalized mutual information gap between the top two latent factors $(I(y_k, z_{I}) - I(y_k, z_{II}))/(\sum_{j=1}^{d} I(y_i, z_j))$. The average MIG score is taken by normalizing the total mutual information. The score reports the compactness of the latent code by ensuring that the information contained in a fixed ground truth $y_k$ is expressed by only one latent factor $z_j$ at a time. High values imply a high level of disentanglement in the latent space. \textbf{Joint Entropy Minus Mutual Information Gap (JEMMIG) score} \citep{do2021theory}. This indirectly measures the modularity in latent space, since a single latent factor may explain more than one ground truth factor. It is expressed as $H(y_k, z_{I}) - I(y_k, z_{I}) + I(y_k, z_{II})$. A lower JEMMIG score is preferred or a high $(1- \text{JEMMMIG})$ score. The average JEMMIG score ($\frac{1}{K}\sum_{k=0}^{K-1}JEMMIG (y_k)$) quantifies the interpretability of the latent variables, measuring both its compactness and its explicitness. \textbf{Modularity score} \citep{ridgeway2018learning} expresses the number of latent factors $z_j$ with high mutual information and explains the ground truth factors.
\comment{
\textbf{Eastwood: Disentanglement (D), Completeness (C) \& Informativeness (I)} \citep{eastwood2018a}. No explicit definition is given for \textbf{disentanglement} score except that a highly disentangled representation $z_i$ with score close to $1$ captures at most one ground-truth factor $y_k$. Similarly, high \textbf{completeness} score close to 1 indicates the information captured by latent $z_i$ about $y_k$ is sufficiently high. An \textbf{informative} representation $z$ scores high if $z$ retains maximum information as regards ground-truth factor $y_k$. Lasso and Random forest regression are both used for mapping to $z_j$'s to $y_k$ by estimating the probabilistic importance of $z_j$ predicting $y_k$.
}

\subsection{Quantitative and Qualitative Evaluation}
Improving disentanglement in latent space without compromising reconstruction, interpretatbility and informativeness is challenging yet achievable. We present a generalized controllable model that automatically adjusts to learning parameters seeking to increase disentanglement, while reducing reconstruction error. We also demonstrate the strength of this model for generative tasks, using well-known data sets and comparing it with benchmark models.
\begin{table*}[htt]
    \centering
    \begin{tabular}{cccccc}
        \hline
         &
          MIG \color{blue}$\uparrow$ &
          Modularity \color{blue}$\uparrow$ &
          JEMMIG \color{blue}$\uparrow$ &
          Reconstruction loss \color{red}$\downarrow$ &
          KL loss \color{blue}$\nearrow$\\ \hline
        VAE         & 0.1268 & 0.798 & 0.233 & 3.339                           & 3.0025  \\
        $\beta$-VAE & 0.0778 & \textbf{0.881} & 0.238 & \textbf{0.012}                           & 35.0295  \\
        ControVAE   & 0.1213 & 0.782 & 0.312 & 0.016                           & 24.3809 \\
        InfoVAE     & 0.1501 & 0.757 & 0.188 & 0.079                           & 10.0621  \\ \hline
        \rowcolor[HTML]{ECF4FF} 
        GCVAE-I     & 0.1507 & 0.844 & 0.236 & \textbf{0.012} & 24.3739 \\
        \rowcolor[HTML]{DAE8FC} 
        GCVAE-II & \textbf{0.2793} & 0.858 & \textbf{0.312} & \textbf{0.012} & \textbf{24.4316} \\
        \rowcolor[HTML]{ECF4FF} 
        GCVAE-III   & 0.1337 & 0.825 & 0.294 & 0.015                           & 24.2937 \\ \hline
    \end{tabular}
    \caption{\label{dist_metric_10}Performance comparison of different models on \textbf{DSprites} after training on 737 samples. Comparison metrics MIG \citep{chen2019isolating}, Modularity \citep{ridgeway2018learning} and JEMMIG \citep{do2021theory} for $10$-D Latent representation. The direction of the arrow indicates the best performing model. Higher is better for MIG, Modularity and JEMMIG($1- JEMMIG$). \textbf{GCVAE-II} performs best on MIG disentanglement metric, robustness and interpretability; plus having the lowest reconstruction error. \textbf{GCVAE-I, III} and \textbf{ControlVAE} also measure up in disentanglement compared to other benchmark models. \textbf{GCVAEs} have the least reconstruction error partly due to the normalization introduced by the inverse precision matrix in $D_{KL}(q_{\phi}(z))||p_{\theta}(z))$ and the weight on the first term of the GCVAE loss \ref{gcvae:eqn1}}
\end{table*}
In the rest of this paper, we focus on information-based metrics like MIG (compactness), JEMMIG (explicitness) and Modularity (robustness) scores, as they are more robust for unsupervised scenarios.
The information based performance metric from Table~\ref{dist_metric_10} measures the quality of disentanglement per model. The MIG metric indicates the best performing model is GCVAE-II and GCVAE-I respectively. 

The GCVAE-II and III both require more computational time per epoch as a result of the approximate cubic time $O({n^3})$ of inverting the precision matrix $\Sigma$. The conditional independence of variables $i, j$ in $z$ noted $\Sigma_{z_{ij}} > 0$ given a third variable increases the precision when estimating the partial correlation of $i, j$ given the rest of the variables. The higher the diagonal covariance, the more stable the approximation of $D_{KL}(q_{\phi}(z))||p_{\theta}(z))$ and robust estimate of $I(y_k, z)$, hence, an $85\%$ increase in estimation of MIG for GCVAE-II compared to I. This represents an advantage over other models, as we now know that a robust estimation of the conditional independence of variables in latent $z$ induces disentanglement.

The reconstruction quality for GCVAE-II is shown to the the least indicating high level of clarity in the reconstruction of the observation.
\subsection{Generation}
\begin{figure*}[ht]
    \centering
    \includegraphics[scale= 0.30]{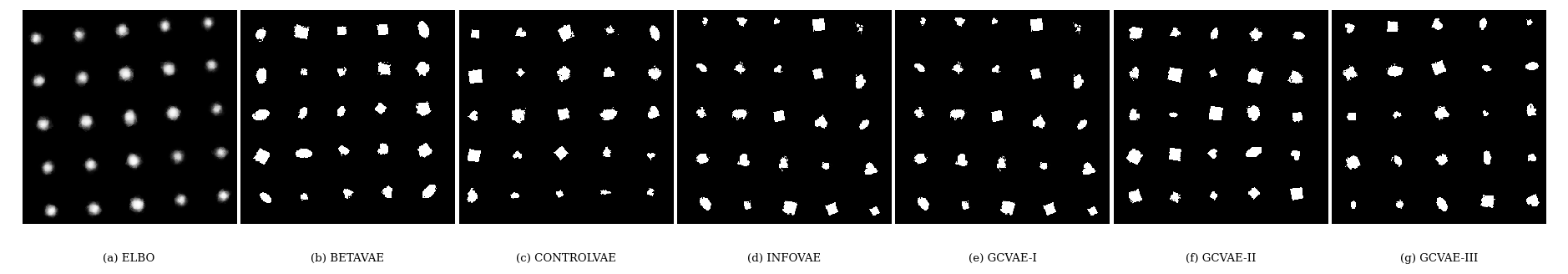}
     \caption{Generative process comparison for the different models on DSprites after training on less than 800 samples. Model GCVAE-I, II and ControlVAE clearly outperformed other models. The reconstruction error of GCVAE-II is the lowest from Table \ref{dist_metric_10}}
\end{figure*}
\begin{figure*}[ht]
    \centering
    \includegraphics[scale= 0.26]{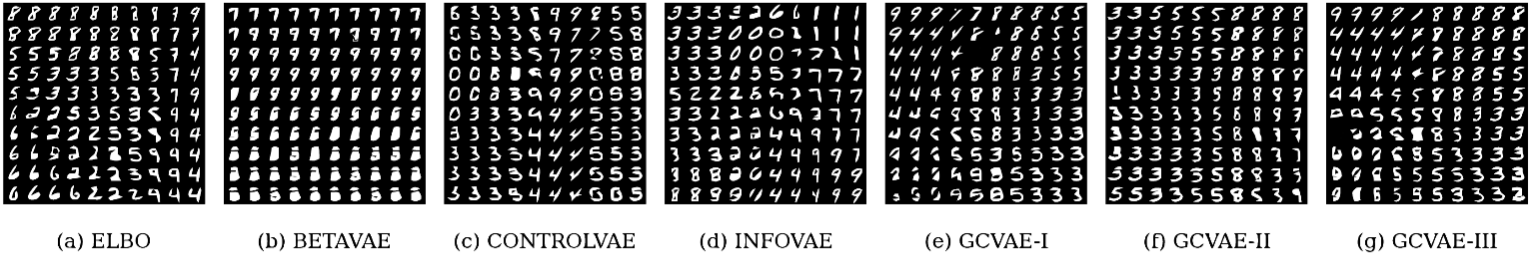}
     \caption{\label{reconst} Generative process comparison for the different models after training the MNIST dataset for $500$ epochs. GCVAE-II and ELBO (VAE) have a similar reconstruction quality with better interpretation. GCVAE-II clearly outperformed the benchmark models in generating clear and meaningful representations of the original data. $\beta$-VAE is the least performing in terms of generating an interpretable image of the original data. ControlVEA and InfoVAE generate meaningful representations.}
\end{figure*}
We evaluate the quality of generation by considering the explicitness and coherency of the encoded latent variables. The generation by GCVAE-I, II and III far outperformed those of the benchmark models (especially on the MNIST dataset). It is worthy of mention that, increasing KL-divergence does not directly correlate to increasing the MIG disentanglement metric (Table \ref{dist_metric_10}). While the Mahalanobis correlation metric is not entirely responsible for disentanglement, it serves as a good heuristic for improving generative quality as well as disentanglement observed in GCVAE-II.

\subsection{Comparing GCVAE-1, II \& III}
\begin{figure*}[ht]
    \centering
    \includegraphics[scale = .25]{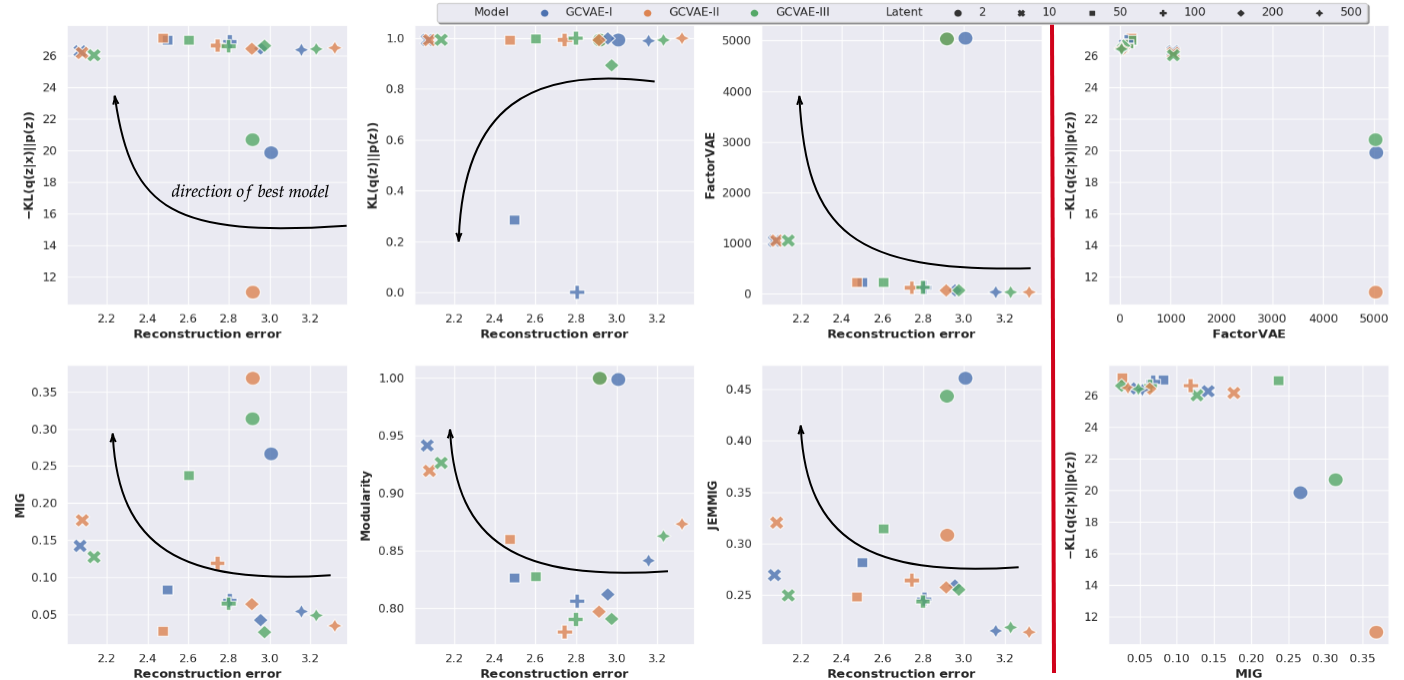}
     \caption{\label{usestop}Model performance comparison on 737 samples of DSprites data. Arrows indicate direction of best performing model. \textbf{Top}: Comparison of reconstruction error against KL divergence , $D_{KL}(q_{\phi}(z|x))||p_{\theta}(z))$ and correlation, $D_{KL}(q_{\phi}(z))||p_{\theta}(z))$. High KL-Low reconstruction error observed for Latent-$10$. Low reconstruction error implies high disentanglement, while high KL does not. High disentanglement using FactorVAE is observed on Latent-$2$ followed by Latent $10$. \textbf{Bottom}: Comparing disentanglement metrics with reconstruction loss. The highest disentanglement on the MIG metric is observed for GCVAE-\textbf{II} on Latent-$2$, however, the best scores are observed for GCVAE-II on Latent-$10$. JEMMIG is similar in behavior to MIG. \textbf{RHS}: Validating the statement \textit{Latent disentanglement is not correlated to KL maximization}.}
\end{figure*}

We compare performance of GCVAEs both on reconstruction and disentanglement using a stopping criteria algorithm. Using a fixed $\epsilon$ for $\alpha_t = 10e-5$ and $\beta_t = 10e-4$, we obtain reasonably low reconstruction loss with high disentanglement without having to train for a lengthy period. The average time required to train GCVAE using the stopping criterion is $6$ hours while it takes more than $3$days to train for 250K iterations. The performance of the model compared for increasing latent dimension is illustrated in Figure \ref{usestop}.

\section{Conclusion}
The challenge of latent space disentanglement for variational models often comes with finding the right reconstruction trade-off. As a result, we propose a new lower bound we refer to as Generalized-Controllable Variational Autoencoder (GCVAE: a model built from a constraint optimization perspective to maximize mutual information in the generative phase, subject to inference constraints to encourage disentanglement in latent space. We use the Mahalanobis distance metric as a heuristic to encourage independence between latent space variables and show that the representation obtained GCVAE is both meaningful and interpretable, with low reconstruction loss. GCVAE-II demonstrates notable strength in disentangling latent space and reconstructing with minimal mutual information loss compared to other variants.

\bibliography{GCVAE_ARXIV}

\clearpage
\appendix
\section{Proof}

\subsection{Maximizing Negative KL-divergence} \label{neg_kl}
Proof that maximizing the negative KL-divergence is equivalent to maximizing the ELBO.
\small
\begin{align}
     \mathcal{L}(\theta, \phi) &= - \underset{p_{\mathcal{D}}}{\mathbb{E}}[D_{KL}(q_{\phi}(z|x)|| p_{\theta}(z|x))] \\
     &= - \underset{p_{\mathcal{D}}}{\mathbb{E}} \underset{z \sim q_{\phi}(z|x)}{\mathbb{E}}\left[\ln \frac{q_{\phi}(z|x)}{p_{\theta}(z|x)}\right]\\
     &= - \underset{p_{\mathcal{D}}}{\mathbb{E}} \underset{z \sim q_{\phi}(z|x)}{\mathbb{E}}\left[\ln \frac{q_{\phi}(z|x)}{p_{\theta}(z)} \cdot \frac{1}{p_{\theta}(x|z)}  \cdot p_{\theta}(x) \right]\\
     \ln p_{\theta}(x) &\geq \underset{p_{\mathcal{D}}}{\mathbb{E}} \left[\ln p_{\theta}(x|z) - D_{KL}(q_{\phi}(z|x)||p_{\theta}(z))\right]
     \label{alt2}
\end{align}
\normalsize

\subsection{Minimizing $I_q(x, z)$} \label{miq}
We demonstrate that minimizing the mutual information between $x$ and latent $z$ equivalently maximizes components of the ELBO and consequently, $I_p(x^{\prime}, z)$ subject to inference constraints.
\small
\begin{align}
    & \underset{\theta, \phi \in \mathbb{R}}{min}  \quad I_q(x, z)
\end{align}
\normalsize

\small

\small
\begin{align}
    I_q(x, z) &= D_{KL}(q_{\phi}(z|x)||q_{\phi}(z))\\
    &= \underset{z \sim q_{\phi}(z|x)}{\mathbb{E}} \left[ \ln \frac{q_{\phi}(z|x)}{q_{\phi}(z)}\cdot \frac{p_{\theta}(x, z)}{p_{\theta}(x, z)}\right] \\
    &= \underset{z \sim q_{\phi}(z|x)}{\mathbb{E}} \left[ \ln \frac{q_{\phi}(z|x)}{p_{\theta}(x|z)p_{\theta}(z)}\cdot \frac{p_{\theta}(z|x)p_{\theta}(x)}{q_{\phi}(z)}\right] \\
    &= \underset{z \sim q_{\phi}(z|x)}{\mathbb{E}} \left[ \ln \frac{q_{\phi}(z|x)}{p_{\theta}(z)}\cdot \frac{p_{\theta}(z|x)}{q_{\phi}(z)} \cdot \frac{p_{\theta}(x)}{p_{\theta}(x|z)}\right] \\
    &= D_{KL}(q_{\phi}(z|x)||p_{\theta}(z)) - D_{KL}(q_{\phi}(z)||p_{\theta}(z)) \notag \\
    &\quad- \underset{z \sim q_{\phi}(z|x)}{\mathbb{E}}[\ln p_{\theta}(x|z)] + \ln p_{\theta}(x)
\end{align}
\normalsize

\small
Note that $p_{\theta}(z|x) \sim p_{\theta}(z)$ so that, 
\small
\begin{align}
    \ln p_{\theta}(x) &\ge \underset{z \sim q_{\phi}(z|x)}{\mathbb{E}}[\ln p_{\theta}(x|z)] -  D_{KL}(q_{\phi}(z|x)||p_{\theta}(z)) \notag\\
        &\quad + D_{KL}(q_{\phi}(z)||p_{\theta}(z)) \label{i_q_min}
\end{align}
\normalsize

\small
Therefore, minimizing $I_q(x, z)$ is equivalent to maximizing terms on the Right Hand Side (R.H.S) of \ref{i_q_min}, which is a lower bound.
\subsection{GCVAE: Proof of proposition} \label{gcvae}
We recall the constraint optimization loss presented in section (\ref{gcvae:sec}) and prove it accordingly. Given the maximization problem,
\small
\begin{align}
    & \underset{\theta, \phi, \xi^{+}, \xi^{-}, \xi_{p} \in \mathbb{R}}{max}\quad I_p(x^{\prime}, z)  \notag\\
    s.t &\quad  \underset{p_{\mathcal{D}}}{\mathbb{E}} D_{KL}(q_{\phi}(z|x)\parallel p_{\theta}(z)) + I_p(x^{\prime}, z)\leq \xi^{-} \notag\\ 
    s.t &\quad - \underset{p_{\mathcal{D}}}{\mathbb{E}} D_{KL}(q_{\phi}(z)||p_{\theta}(z)) \leq \xi^{+} \notag\\
    s.t &\quad I_p(x^{\prime}, z) \leq \xi_{p} \notag\\
    s.t &\quad \xi_i^{+}, \xi_i^{-}, \xi_{ip} \geq 0, \quad \forall i = 1, \dots, n
\end{align}
\normalsize

\small
The expand of the above equations using sets of Lagrangian multipliers is as follows,
\small
\begin{align}
    &\mathcal{L}(x, z; \theta, \phi, \xi^{+}, \xi^{-}, \xi, \alpha, \beta, \gamma, \bm{\eta}, \bm{\tau}, \bm{\nu}) \qquad \qquad \qquad\qquad \notag\\
    &= I_p(x^{\prime}, z) - \beta(\underset{p_{\mathcal{D}}}{\mathbb{E}} D_{KL}(q_{\phi}(z|x)|| p_{\theta}(z)) + I_p(x^{\prime}, z) - \sum_{i = 1}^{n}\xi_i^{-}) \notag\\
    &+ \gamma(\underset{p_{\mathcal{D}}}{\mathbb{E}} D_{KL}(q_{\phi}(z)||p_{\theta}(z)) + \sum_{i = 1}^{n}\xi_i^{+}) - \alpha (I(x^{\prime}; z) - \sum_{i = 1}^{n}\xi_{ip}) \notag \\
    &+ \sum_{i = 1}^{n} \bm{\eta}_i \xi_{i}^{+} + \sum_{i = 1}^{n} \bm{\tau}_i \xi_{i}^{-} + \sum_{i = 1}^{n} \bm{\nu}_i \xi_{ip}
    \label{dcvae:eqn3}
\end{align}
\normalsize

\small
\begin{align}
    \mathcal{L}(& x, z; \theta, \phi, \xi^{+}, \xi^{-}, \xi, \alpha, \beta, \gamma, \bm{\eta}, \bm{\tau}, \bm{\nu}) \qquad \qquad \qquad\qquad \notag\\
    &= (1- \alpha- \beta)I_p(x^{\prime}, z) - \beta\underset{p_{\mathcal{D}}}{\mathbb{E}} D_{KL}(q_{\phi}(z|x)|| p_{\theta}(z)) \notag\\
    &+ \gamma D_{KL}(q_{\phi}(z)||p_{\theta}(z)) + (\beta - \bm{\tau})\sum_{i = 1}^{n}\xi_i^{-}  + (\gamma - \bm{\eta})\sum_{i = 1}^{n}\xi_i^{+} \notag \\
    &+ (\alpha - \bm{\nu})\sum_{i = 1}^{n}\xi_{ip}
    \label{dcvae:eqn4}
\end{align}
\normalsize

We take the gradient over the loss, $\triangledown \mathcal{L}$ for $\xi^{-}, \xi^{+}, \xi_{p}$ and apply KKT optimality conditions to obtain,
\small
\begin{align}
    \mathcal{L}(& x, z; \theta, \phi, \bm{\xi^{+}}, \bm{\xi^{-}}, \bm{\xi_{p}}, \alpha, \beta, \gamma) \qquad \qquad \qquad\qquad \notag\\
    &= (1- \alpha- \beta)I_p(x^{\prime}, z) - \beta\underset{p_{\mathcal{D}}}{\mathbb{E}} D_{KL}(q_{\phi}(z|x)||p_{\theta}(z)) \notag\\
    &\quad+ \gamma D_{KL}(q_{\phi}(z)||p_{\theta}(z))\\
    &= (1- \alpha- \beta)\underset{z \sim q_{\phi}(z|x)}{\mathbb{E}}[\ln p_{\theta}(x|z)] \notag\\
    &\quad- \beta\underset{p_{\mathcal{D}}}{\mathbb{E}} D_{KL}(q_{\phi}(z|x)|| p_{\theta}(z)) \notag\\
    &\quad+ \gamma D_{KL}(q_{\phi}(z)||p_{\theta}(z))
    \label{dcvae:eqn5}
\end{align}
\normalsize

We set the Lagrangian adaptive hyperparameters following \citep{controlvae} as follows,
\small
\begin{align}
    \mathcal{L}(x, z; \theta, \phi, \alpha, &\beta, \gamma) \qquad \qquad \qquad\qquad \notag\\
    &= (1- \alpha_t- \beta_t)\underset{z \sim q_{\phi}(z|x)}{\mathbb{E}}[\ln p_{\theta}(x|z)] \notag\\
    &\quad- \beta_t\underset{p_{\mathcal{D}}}{\mathbb{E}} D_{KL}(q_{\phi}(z|x)|| p_{\theta}(z)) \notag\\
    &\quad+ \gamma_t D_{KL}(q_{\phi}(z)||p_{\theta}(z))
    \label{dcvae:eqn6}
\end{align}
\normalsize

The adaptive weight $\alpha_t$ controls the reconstruction error while $\beta_t$ ensures the posterior latent factor $q_{\phi}(z|x)$ does not deviate significantly from its prior $p_{\theta}(z)$. Varying both terms gives us better control of the degree of disentanglement and helps us to understand the parameters affecting density disentanglement.

\subsection{Expected Squared Mahalanobis distance: $\mathbb{E} D_{MAH}^{2}(q||p)$} \label{mmd_mal_prove}
Suppose that a data $x \in \mathcal{X}$ with probability $p(x)$ is projected into a reproducing kernel Hilbert space $<\varphi (x)>_{\mathcal{H}}$, the expectation of the squared Mahalanobis distance is as follows,

\begin{figure*}[ht]
    \centering
    \includegraphics[scale= 0.30]{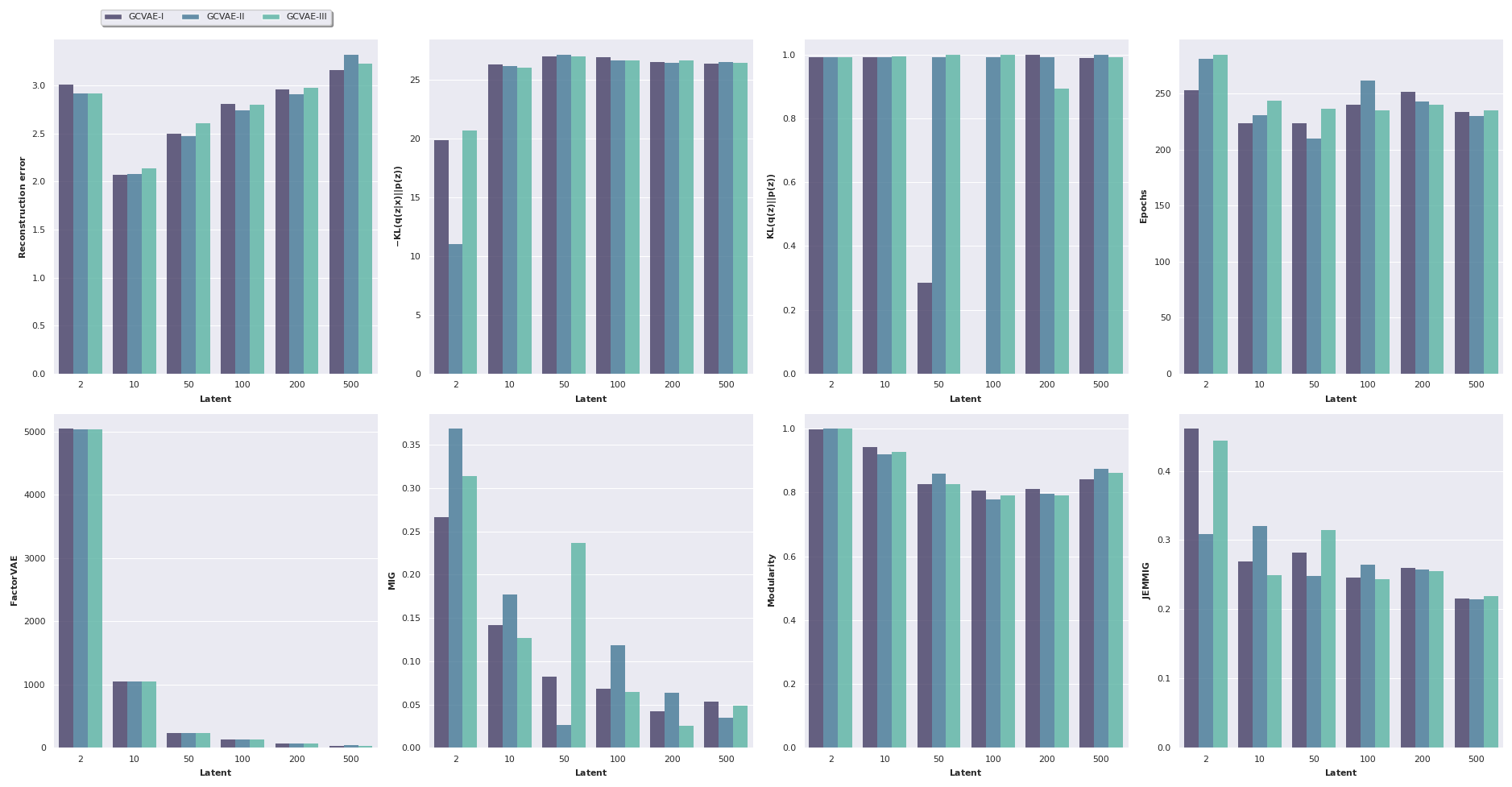}
     \caption{ \label{gc_sc} Comparison of metrics for DSprites 2D shapes dataset. \textbf{Top}: Comparison of GCVAE-I, II \& III losses over increasing latent dimensions. The lowest reconstruction error is observed for GCVAE-I on Latent-$10$ and is monotone increasing thereafter. $-D_{KL}(q_{\phi}(z|x))||p_{\theta}(z))$ is stable, increasing above $z=10$ dimensions and highest for GCVAE-II and III on Latent-50. Note that the original values of $D_{KL}(q_{\phi}(z))||p_{\theta}(z))$ are scaled to $[0, 1]$.\\
     \textbf{Bottom}: Disentanglement metric over different dimensions. FactorVAE metric is monotone decreasing for latent space greater than $2$. MIG is similar in behavior to FactorVAE metric and best for GCVAE-II on Latent-$2$. }
\end{figure*}

\small
\begin{align}
    \mathbb{E} & D_{MAH}^{2} (q(z)||p(z)) \qquad\qquad\qquad\qquad\qquad\qquad\qquad\qquad\qquad\qquad\qquad\qquad\notag\\
    &\geq \parallel \underset{x\sim p}{\mathbb{E}}\varphi(x) - \underset{y\sim q}{\mathbb{E}}\varphi(y) \parallel^{2}_{\Sigma^{-1}, \mathcal{H}}\\
    &\geq (\underset{x\sim p}{\mathbb{E}}\varphi(x) - \underset{y\sim q}{\mathbb{E}}\varphi(y))^{\prime}\Sigma^{-1}(\underset{x\sim p}{\mathbb{E}}\varphi(x) - \underset{y\sim q}{\mathbb{E}}\varphi(y))\\
    &\geq \Sigma^{-1}(\underset{x\sim p}{\mathbb{E}}\varphi(x) - \underset{y\sim q}{\mathbb{E}}\varphi(y))^{\prime}(\underset{x\sim p}{\mathbb{E}}\varphi(x) - \underset{y\sim q}{\mathbb{E}}\varphi(y)) \\
    &\geq \Sigma^{-1}[\underset{x^{\prime}\sim p, x\sim p}{\mathbb{E}} \langle\varphi(x^{\prime}), \varphi(x)\rangle_{\mathcal{H}} - \underset{x^{\prime}\sim p, y\sim q}{2\mathbb{E}}\langle\varphi(x^{\prime}), \varphi(y)\rangle_{\mathcal{H}} \notag\\
    &\quad + \underset{y^{\prime}\sim q, y\sim q}{\mathbb{E}}\langle\varphi(y^{\prime}), \varphi(y)\rangle_{\mathcal{H}} ]
\end{align}
\normalsize

We suppose that the feature map $\varphi(x)$ takes the canonical form $k(x, \cdot)$, so that $\langle\varphi(x^{\prime}), \varphi(x)\rangle_{\mathcal{H}} = k(x^{\prime}, x)$ \citep{book_svm}, where $k(x^{\prime}, x)$ represents a positive semi-definite kernel (for instance the Gaussian kernel, $exp(-\frac{\parallel x^{\prime} - x\parallel^{2}}{2\sigma^2})$).
Hence,
\small
\begin{align}
    \mathbb{E} & D_{MAH}^{2} (q(z)||p(z)) \qquad\qquad\qquad\qquad\qquad\qquad\qquad\qquad\qquad\qquad\qquad\qquad\notag\\
    &\geq \Sigma^{-1}[\underset{x^{\prime}\sim p, x\sim p}{\mathbb{E}} k(x^{\prime}, x) - 2\underset{x^{\prime}\sim p, y\sim q}{\mathbb{E}}k(x^{\prime}, y) \notag\\
    &\quad + \underset{y^{\prime}\sim q, y\sim q}{\mathbb{E}}k(y^{\prime}, y) ] \notag\\
    &\geq \Sigma^{-1}[\underset{z^{\prime}\sim p, z\sim p}{\mathbb{E}} k(z^{\prime}, z) - 2\underset{z^{\prime}\sim p, z\sim q}{\mathbb{E}}k(z^{\prime}, z) \notag\\
    &\quad + \underset{z^{\prime}\sim q, z\sim q}{\mathbb{E}}k(z^{\prime}, z) ] \notag\\
    &= \Sigma^{-1} D_{MMD}^{2}(q(z)||p(z))\notag\\
\end{align}
\normalsize

\section{Performance comparison of GCVAE}
We compare the performance of GCVAE-I, II \& III using a stopping criterion. We design a stopping criterion using the adaptive convergence of $\alpha$ and $\beta$ below a certain threshold $\epsilon$. $\epsilon_a$ for stopping reconstruction loss and  $\epsilon_b$ stops KL divergence, $-D_{KL}(q_{\phi}(z|x))||p_{\theta}(z))$ at the optimum point.

\begin{algorithm}[H]
		\SetKwInOut{Input}{Input}
		\SetKwInOut{Output}{Output}
		\caption{GCVAE stopping criteria}
		\Input{$\mathcal{L}(\phi, \theta, \dots), \alpha_t, \beta_t, \gamma_t$, $\epsilon_a$, $\epsilon_b$}
		\Output{Optimal parameters of loss $\mathcal{L}$}
		\Begin{
			\textbf{Compute gradient:} $\nabla \mathcal{L} (\phi, \theta, \alpha_t, \beta_t, \gamma_t, \dots)$\;\hfil
			\textbf{Perform back propagation}\;\hfil
			\If{$\alpha_t$ - $\alpha_{t-1}$ $<$ $\epsilon_a$ $\&\&$ $\beta_t$ - $\beta_{t-1}$ $<$ $\epsilon_b$ }{
			return optimal $\mathcal{L}, \alpha_t, \beta_t, \gamma_t$\;
			}
		} 
\end{algorithm}
The stopping criteria reduces the constraint of stopping the algorithm at a specific learning epoch, serving to assist the model to conclude the final epoch, when the optimal reconstruction and KL divergences are obtained. The optimal parameters yield the best disentanglement with low reconstruction error. We observe that reconstruction error is not usually as low as when run for N-iterations, but the representations results are reasonably comparable. 

Figure \ref{gc_sc} illustrates the result of using the stopping criterion on the DSprites 2D data set. We observe in the very top right-hand corner of this figure that the number of iterations required to reach optimum point is below $300$ for all latent spaces. The singular downside to using stopping criteria is with regards to reconstruction error. Minimum reconstruction error and maximum disentanglement are observed for GCVAE-III on Latent-$10$ and GCVAE-II on Latent-$2$ respectively.

\begin{figure}[ht]
    \centering
    \includegraphics[scale= 0.47]{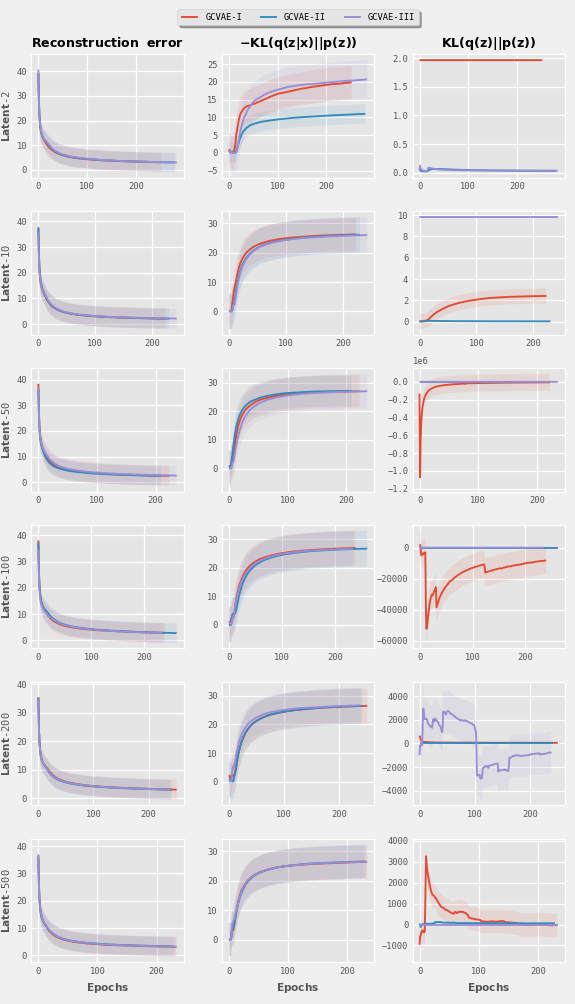} 
    \caption{\label{gc_gc}A visual comparison of the reconstruction, $-D_{KL}(q_{\phi}(z|x))||p_{\theta}(z))$ and $D_{KL}(q_{\phi}(z))||p_{\theta}(z))$ losses for GCVAE-I, II \& III over different latent space. Behavior of reconstruction error per latent is relatively close and indistinguishable. In all cases of latents experimented with except for Latent-2,  $-D_{KL}(q_{\phi}(z|x))||p_{\theta}(z))$ is comparable. GCVAE-I is unstable in $D_{KL}(q_{\phi}(z))||p_{\theta}(z))$ during training across all latents. While a lower value of $D_{KL}(q_{\phi}(z))||p_{\theta}(z))$ is preferred, we observe correlation with MIG disentanglement metric in Figure \ref{gc_sc}.}
\end{figure}

We present an elaborate representation of the results of GCVAE for N-number of iterations in Figure \ref{gc_gc}. The running time is better represented in Figure \ref{gc_sc} while \ref{gc_gc} details reasons for why a GCVAE model obtains a lower MIG disentanglement metric over the other. This is evident in $D_{KL}(q_{\phi}(z))||p_{\theta}(z))$. We observe a collapse of the $D_{KL}(q_{\phi}(z))||p_{\theta}(z))$ when $D_{KL}(q_{\phi}(z))||p_{\theta}(z)) = 0$, which reduces the GCVAE model to ControlVAE. This is demonstrable, since $\alpha_t$ and $\beta_t$ are usually very low. This effect corresponds to the lowest MIG disentanglement metric observed in Latent-$50$ and $200$ for GCVAE-II. A GCVAE model whose posterior $q_{\phi}(z)) \approx p_{\theta}(z)$ returns better disentanglement, since the second $D_{KL}(q_{\phi}(z|x))||p_{\theta}(z)) > 0$.

Secondly, we identify another reason for the collapse of the correlation term, $D_{KL}(q_{\phi}(z))||p_{\theta}(z))$ . Given an inverse covariance matrix ($\Sigma^{-1}$) whose diagonals are infinitesimally small, this could in fact collapse the correlation and subsequently return low disentanglement score. This is indeed evident, since the generated factors of the latent $z$ have a different posterior, $p_{\theta}(x|z)$ from the original factors of the given distribution.

Despite the emphasis on disentanglement as opposed to reconstruction, we observe that GCVAE with a stopping criteria running for very low number of iterations is capable of achieving reasonable reconstruction of the original data as well as generating meaningful data from an unseen sample.

\begin{figure}[ht]
    \centering
    \includegraphics[scale= 0.6]{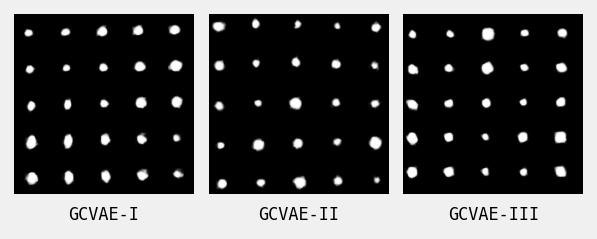}
    \caption{\label{gc_gd}Generated samples from training with 737 observations and stopping after less than 300 iterations. The generated samples especially for GCVAE-II.}
\end{figure}

\subsection{Architecture \label{archi}}
We use similar architecture to train GCVAE-I, II, III and all benchmark models. The proposed convolutional 2D convolutional architecture is given in the table below,

\begin{table}[htt]
\footnotesize
\begin{tabular}{|l|l|}
\hline
Encoder                       & Decoder                         \\ \hline
Input 64 x 64 binary image    & Input $\mathbb{R}^{10}$                       \\ \hline
4 × 4 conv. 64 ReLU. stride 2 & FC 25; FC 200 ReLU              \\ \hline
4 × 4 conv. 64 ReLU. stride 2 & 4 × 4 upconv. 32 ReLU. stride 2 \\ \hline
4 × 4 conv. 32 ReLU. stride 2 & 4 × 4 upconv. 32 ReLU. stride 2 \\ \hline
4 × 4 conv. 32 ReLU. stride 2 & 4 × 4 upconv. 64 ReLU. stride 2 \\ \hline
FC 200; FC 25; FC 2 X 10      & 4 × 4 upconv. 64 ReLU. stride 2 \\ \hline
                              & 4 × 4 upconv. 1. stride 2       \\ \hline
\end{tabular}
\caption{Encoder-Decoder architecture for 2D DSprites and MNIST data. We use a 2D MaxPooling and Batch normalization for the encoder while size $2$ Upsampling with Batch normalization is used for the decoder.}
\end{table}

\end{document}